# Adaptive Shooting for Bots in First Person Shooter Games Using Reinforcement Learning

Frank G. Glavin and Michael G. Madden

*Abstract*—In current state-of-the-art commercial first person shooter games, computer controlled bots, also known as nonplayer characters, can often be easily distinguishable from those controlled by humans. Tell-tale signs such as failed navigation, "sixth sense" knowledge of human players' whereabouts and deterministic, scripted behaviors are some of the causes of this. We propose, however, that one of the biggest indicators of nonhuman-like behavior in these games can be found in the weapon shooting capability of the bot. Consistently perfect accuracy and "locking on" to opponents in their visual field from any distance are indicative capabilities of bots that are not found in human players. Traditionally, the bot is handicapped in some way with either a timed reaction delay or a random perturbation to its aim, which doesn't adapt or improve its technique over time. We hypothesize that enabling the bot to learn the skill of shooting through trial and error, in the same way a human player learns, will lead to greater variation in game-play and produce less predictable nonplayer characters. This paper describes a reinforcement learning shooting mechanism for adapting shooting over time based on a dynamic reward signal from the amount of damage caused to opponents.

## I. Introduction

### A. First Person Shooter Games

THE FIRST PERSON SHOOTER (FPS) genre of computer games has existed for over twenty years and involves a human player taking control of a character, or avatar, in a complex 3D world and engaging in combat with other players, both human and computer-controlled. Human players perceive the world from the first person perspective of their avatar and must traverse the map, collecting health items and guns, in order to find and eliminate their opponents. The most straightforward FPS game type is called a 'Death Match' where each player must work by themselves with the objective of killing more opponents than anyone else. The game ends when the score limit has been reached or the game time limit has elapsed. An extension to this game type, 'Team Death Match', involves two or more teams of players working against each other to accumulate the most kills. Objective based games also exist where the emphasis is no longer on kills and deaths but on specific tasks in the game which, when successfully completed, result in acquiring points for your team. Two examples of such games are 'Capture the Flag' and 'Domination'. The former involves retrieving a flag from the enemies' base and returning it to your base without dying. The latter involves keeping control of predefined areas on the map for as long as possible. All of these game types require, first and foremost, that the player is proficient when it comes to combat. Human players require many hours of practice in order to become familiar with the game controls and maps and to build up quick reflexes and accuracy. Replicating such human behavior in computer-controlled bots is certainly a difficult task and it is only in recent years that gradual progress has been made, using various artificial intelligence algorithms, to work towards accomplishing this task. Some of these approaches will be discussed later in Section II.

### B. Reinforcement Learning

Reinforcement learning [1] involves an *agent* interacting with an *environment* in order to achieve an explicit goal or goals. A finite set of states exist, called the *state space*, and the agent must choose an available action from the *action space* when in a given state at each time step. The approach is inspired by the process by which humans learn. The agent learns from its interactions with the environment, receiving feedback for its actions in the form of numerical rewards, and aims to maximize the reward values that it receives over time. This process is illustrated in Fig. 1. The state-action pairs that store the expected value of carrying out an action in a given state comprise the *policy* of the learner. The agent must make a tradeoff between exploring new actions and exploiting the knowledge that it has built up over time. Two common approaches to storing/representing policies in reinforcement learning are *generalization* and *tabular*. With generalization, a function approximator is used to generalize a mapping of states to actions. The tabular approach, which is used in this research, stores numerical representations of all state-action pairs in a lookup table. The specific policy-learning algorithm that we use in this work is Sarsa($\lambda$), which will be described later in Section III.

### C. Problem Summary

FPS games require human players to have quick responses, good hand-eye coordination and the ability to memorize complex game controls. In addition to this, they must also remember the pros and cons of specific guns, learn the layout of the different maps and develop their own unique playing style that works well for them. Some players prefer an aggressive "run

Manuscript received January 07, 2014; revised May 29, 2014; accepted October 10, 2014. Date of publication October 14, 2014; date of current version June 12, 2015. This work was supported by a Scholarship funded through the Higher Education Authority of Ireland.

The authors are with the College of Engineering and Informatics, National University of Ireland, Galway, Ireland (e-mail: frank.glavin@nuigalway.ie; michael.madden@nuigalway.ie).

Color versions of one or more of the figures in this paper are available online at http://ieeexplore.ieee.org.

Digital Object Identifier 10.1109/TCIAIG.2014.2363042

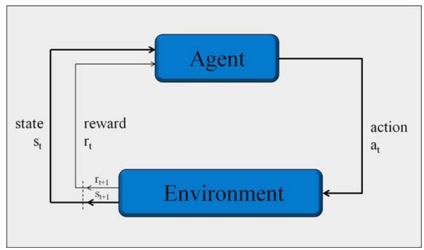

Fig. 1. The interactions between the agent and the environment (This figure is based on [1 , Fig. 3.1 ]).

and gun" approach while others are more reserved and cautious while playing. It is this diversity that leads to interesting and entertaining gameplay where players build up experience and find new ways to out-wit their opponents. Artificially generating such skills and traits in a computer controlled bot is a difficult and complex task. While bots can be programmed relatively easily to flawlessly carry out the tasks required to play in a FPS, this is not the goal of developing effective AI opposition in these games. The overall aim is to design a bot in such a way that a human player would not be able to detect that the opponent is being controlled by a computer. In this sense, bots cannot be perfect and must be occasionally prone to bad decision making and "human errors" while at the same time learning from their mistakes. In our previously published work [2], [3], we developed a general purpose bot that used reinforcement learning with multiple sources of reward. In this research, we are only concerned with the development and adaptation of shooting skills over time. This is just one task of many that we believe will lead to creating entertaining and human-like NPCs in the future. The shooting architecture can be "plugged in" to existing bots, overriding the default shooting mechanism. We believe it is important to develop and analyze each task individually before merging them together into the final bot version. Examples of other tasks would include navigation, item collection, and opponent evasion.

## II. Related Research

Reinforcement learning has been used to embed game AI in many different genres of computer games in the past. These genres include Real Time Strategy (RTS) games [4]–[6], fighting games [7], [8], and board games [9]–[11] among others. Improving NPC behaviors in FPS games has also received noteable attention with everincreasing PC peformance and the advent of "next generation" gaming consoles. This section examines some of the artificial intelligence approaches used in FPS games that are related to this research.

In 2008, a competition was set up for testing the humanness of computer controlled bots in FPS games. This was called BotPrize [12] and the original competition took place as part of the IEEE Symposium on Computational Intelligence and Games[1]. The purpose of the competition, which has been repeated annually, is to see whether computer-controlled bots can fool human observers into thinking that they are human players in the FPS game Unreal Tournament 2004 (UT2004). In this sense, the competition essentially acts as a Turing Test [13] for bots. Under the terms of the competition, a bot is successful if it fools observers into believing that it is human at least 50% of the time. The original design of the competition involved a judge playing against two opponents (one human and one bot) in 10–min death matches. The judge would then rank the players on a scale of 1 to 5 in humanness. The improved design [14] made the judging process part of the game. An existing weapon in UT2004 called the Link Gun was modified and this is used to judge other players as being humans or bots. The damage caused by each gun in the competition is set at 40% of the normal damage, to give players enough time to make an informed decision. This competition ran for five years before finally being won by two teams in 2012. MirrorBot (52.2%) and the $UT^2$ bot (51.9%) surpassed the "humanness barrier" of 50%.

MirrorBot, developed by Polceanu [15], records opponents' movements in real time and if it encounters what it perceives to be a nonviolent player it will trigger a special behavior of *mirroring*. The bot then proceeds to mimic the opponent by playing back the recorded actions after a short delay. The actions are not played back exactly as recorded to give the impression that they being independently selected by the bot. MirrorBot has an aiming module to adjust the bot's orientation to a given focus location. If the opponent is moving then a "future" location will be calculated based on the opponents velocity and this will be used to target the opponent. In the absence of a target, MirrorBot will focus on a point computed from a linear interpolation of the next two navigation points. The authors do not report any weapon-specific aiming so it is assumed that this aiming module is used for all guns despite the large variance in how different ones shoot. The decision on which weapon to use is based on its efficiency and the amount of available ammunition.

The $UT^2$ bot, developed by Schrum *et al.* [16] uses human trace data when it detects that it is stuck (navigation has failed). The authors also developed a combat controller using neuroevolution which evolves artificial neural networks, where the fitness function is designed to encourage human-like traits in game play. For its shooting strategy, the bot shoots at the location of the opponent with some random added noise. The amount of noise added is dependent on the distance from the opponent and its relative velocity with more noise being added as the distance and relative velocity values increase. Full development details and an analysis of the bot's performance in Bot Prize can be found in the chapter by Schrum *et al.* [17]

Gamez *et al.* [18] developed a system which uses a global workspace architecture implemented in spiking neurons to control a bot in Unreal Tournament 2004. The system is designed to create a bot that produces human-like behavior and the architecture is based on control circuit theories of the brain. It is the first system of this type to be deployed in a dynamic realtime environment. The bot was specifically designed to reproduce humanlike behavior and competed in the Bot Prize competition in 2011, coming in second place with a humanness rating of 36%. The authors also developed a metric for measuring the humanness of an avatar by combining a number of statistical measures into a single value. These were exploration factor, stationary time, path entropy, average health, number of kills and number of deaths. The exploration factor metric measures how much of the available space on the map is visited by the avatar. Stationary time measures the total amount of time that the avatar is stationary during the game. Path entropy measures variability in the avatars movements while navigating. The humanness metric is calculated as the average of all of these statistical measures. Using this humanness metric, the authors found that a similar humanness rating percentage was obtained to those that were calculated through the use of human judges in the Bot Prize competition. The authors do not report any implemented variance in the shooting action of the bot.

[1]http://www.botprize.org/2008.

McPartland and Gallagher [19] applied the tabular Sarsa($\lambda$) reinforcement learning algorithm to a simplified purpose-built first person shooter game. Individual controllers were trained for navigating the map, collecting items and engaging in combat. The experimentation involved three variations of the reinforcement learning algorithm. The first of these, *HierarchicalRL*, learns when to use the combat or navigation controller. *RuleBasedRL* has predetermined rules for deciding on which controller to use and the *RL* controller which learns the entire task of navigation and combat from scratch. A comparative analysis was carried out which included a *random* bot and *state machine* bot. The results showed that the reinforcement learning bots performed well in this purpose-built FPS game. McPartland and Gallagher [20] extended this research by developing an interactive training tool in which human users can direct the policy of the learning algorithm. The bot follows its own policy unless otherwise directed by the user. They also investigated the outcome of having five commercial game developers use the interactive tool to train bots [21]. They concluded from their experiments that the training could produce bots with different behavior styles in the simplified environment. The developers reported that the training tool had potential for use in FPS game development and they also identified several improvements that could be made. Our work differs from that of McPartland and Gallagher in several ways. First, we have developed an architecture for shooting which is embedded in a commercial FPS game as opposed to a simplified, purpose-built one. Second, we have tailored the architecture to be "plugged into" an existing game to replace some of the core functionality of the bots logic, in this case, how it shoots. The reward signal used is also dynamic and taken directly from the systems reporting of damage caused to opponents. Finally, the states and actions have been designed from the perspective of a human player playing an FPS game and "snapshots" of the bots memory are stored as the bot learns. Any stage of the bots learning can be loaded at the beginning of a new game.

Tastan *et al.* [22] developed an Unreal Tournament bot that uses maximum entropy inverse reinforcement learning and particle filtering for the problem of opponent interception. First, human trace data is used to learn a reward function which can then generate a set of potential paths that the opponent could be following. These learned paths are then maintained as hypotheses in a particle filter. Separate particle layers are run for tracking probable locations at different times. The final step involves planning a path for the bot to follow.

Conroy *et al.* [23] carried out a study to analyse human players responses to computer-controlled opponents in FPS

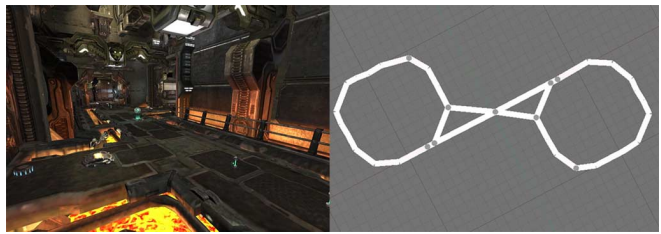

Fig. 2. Training Day map and a birds eye view of its layout.

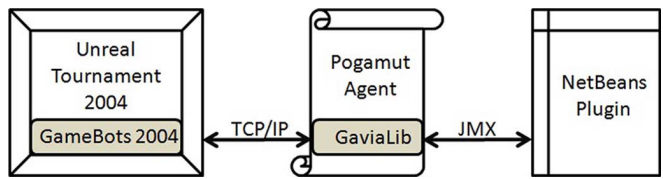

Fig. 3. Pogamut 3 Architecture (based on Gemrot *et al.*, [24, Fig. 1]).

games. The study examined how well the players can distinguish between other humans and NPCs while also seeking to identify some of the characteristics that lead to an opponent being labelled as artificially controlled. A multiplayer game play session was carried out with 20 participants in Quake III followed by a survey. The top opponent behaviors used by survey takers for making judgements were aiming, camping (lying in wait for the opponent), response to player, and fleeing from combat.

## III. METHODOLOGY

### A. Unreal Tournament 2004 and Pogamut 3

The reinforcement learning shooting architecture for this research was developed using the game UT2004 and an opensource development toolkit called Pogamut 3.

UT2004 is a first person shooter game, and the third game released under the Unreal franchise, developed primarily by Epic Games[2]. It has 11 different game types, including those mentioned in Section I-A, and the game also includes modding capabilities with many user-made maps and player models which can be found online. There is also an extensive array of weapons available with 19 of them in total. The weapons available in each game depend on the map being played. There are points on each map where different guns appear as pick-ups. The map shown in Fig. 2 is called Training Day. This is one of the smallest maps in the game and is used for our experimentation in Section IV. UT2004 uses the Unreal Engine which has a scripting language called UnrealScript for high-level programming of the game. Players can compete against other human players online as well as being able to play against computer-controlled bots, or a combination of both humans and bots.

Pogamut 3 [24] facilitates the creation of bots for UT2004. It has modules that simplify the process of adding capabilities for the bot in the game, such as navigation and item collection, so that development work can be focused on the artificial intelligence which drives the bots' behavior. Pogamut 3 integrates five main components: UT2004, GameBots2004, the GaviaLib Library, the Pogamut agent and the NetBeans plugin Integrated Development Environment (IDE). This is illustrated in Fig. 3. GameBots2004, an extension to the original GameBots [25], uses a TCP/IP text-based protocol so users can connect their agents to UT2004 in a client-server architecture where GameBots2004 acts as the server. The GaviaLib library is a Java library that acts as an interface for accessing virtual environments such as UT2004. The agent interface that it provides comprises classes for listening for events and querying object instances. The agent itself is made up of Java classes and interfaces which are derived from the classes of the GaviaLib library. The IDE is a NetBeans plugin that communicates with the agent using JMX. The IDE includes project templates, example agents, server management, access to agent properties and a log viewer among other features. A fully detailed description of this architecture can be found in Gemrot et al. [24].

---

[2]http://liandri.beyondunreal.com/Unreal_Tournament_2004.

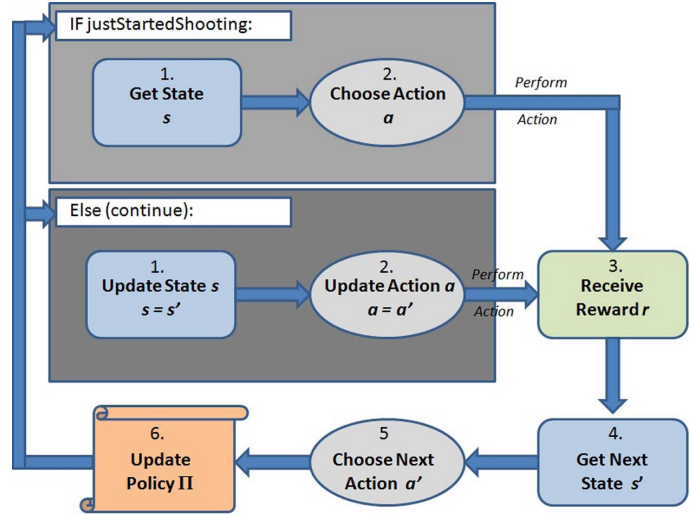

Fig. 4. Bot shooting logic using Sarsa($\lambda$).

B. Sarsa($\lambda$) Algorithm

Tabular Sarsa($\lambda$) [1] is an on-policy algorithm which involves an agent interacting with the environment and updating its policy based on the actions that are taken. At the beginning of a new episode, the current state is obtained and then an action is selected from all available actions in this state, based on some action-selection policy. These policies are nondeterministic and involve some amount of exploration in the policy. The purpose of these policies is to balance the tradeoff between exploring new actions and exploiting the knowledge that has already been learned. The $\epsilon$-greedy action-selection policy is used in this research. With this approach, the most favorable action is chosen $1 - \epsilon$ of the time from those available (i.e., the one with the highest estimated Q-value recorded so far) but a random action is performed $\epsilon$ of the time. For example, if $\epsilon$ is set to 0.3 then a random action will be chosen 30% of the time. Random actions are chosen with a uniform probability distribution. The algorithm uses eligibility traces to speed up learning by allowing past actions to benefit from the current reward. The use of eligibility traces can enable the algorithm to learn sequences of actions, which could be useful when learning effective shooting strategies in FPS games. The pseudocode for the algorithm is presented in Algorithm 1.

**Algorithm 1 Pseudocode for the Sarsa($\lambda$) algorithm.**

    **for all** $s, a$ **do**

        $Q(s, a) = 0$

        $e(s, a) = 0$

    **end for**

    **repeat**

        Initialize $s, a$

        **repeat**

            Take action $a$, observe $r, s'$

            Choose $a'$ and $s'$ using policy derived from $Q$

            $\delta \Leftarrow r + \gamma Q(s', a') - Q(s, a)$

            $e(s, a) \Leftarrow 1$

            **for all** $s, a$

                $Q(s, a) \Leftarrow Q(s, a) + \alpha \delta e(s, a)$

                $e(s, a) \Leftarrow \gamma \lambda e(s, a)$

            **end for**

            $s \Leftarrow s'; a \Leftarrow a'$

        **until** (steps of single episode have finished)

    **until** (all episodes have finished)

For the experiments reported in this paper, we use the following values for the Sarsa($\lambda$) parameters. The learning rate $\alpha$ determines how quickly newer information will override older information that was learned. As the value approaches 1, the agent will only consider the most recent information. If the value was closer to 0, then the current information would have less of an immediate impact on the learning. We would like the bot to have strong consideration for recent information without completely overriding what has been learned so the value is set to 0.7. The discount parameter $\gamma$ determines how important future rewards are. The closer the value is to 0, the more the agent will only consider current rewards whereas a value close to 1 would mean the agent would be more focused on long term rewards. To enable a balance between current and longterm rewards we set $\gamma$ to 0.5. The eligibility trace, $\lambda$, is set to 0.9. This value represents the rate at which the eligibility traces decay over time. This large value results in recent state-action pairs receiving a large portion of the current reward.

The algorithm works as follows. First, the Q-values, $Q(s,a)$, and eligibility traces, $e(s,a)$, for all states and actions are initialized to 0. At the beginning of each episode, the current state and current action values, $s$ and $a$, are initialized. Then, for every step of each episode, the action $a$ is taken and a reward $r$ is received, and the next state $s'$ is observed. The next action $a'$ is then chosen from the next state using the policy ($\epsilon$-greedy). The temporal difference (TD) error $\delta$ is then calculated using $r$, $\gamma$ and the current and next state-action pairs. TD learning uses principles from Monte Carlo methods and Dynamic Programming (DP) in that it learns from an environment based on a policy and it approximates its estimates based on previously learned estimates (this is known as bootstrapping [1]). The current eligibility trace is then assigned a value of 1 to mark it as being eligible for learning. Next, the Q-values and eligibility traces for all states and actions are updated as follows. Each Q-value is updated as the old Q-value plus the eligibility trace variable multiplied by the learning rate and the TD error. Each eligibility trace variable is then updated as the old value multiplied by the discount parameter and the eligibility trace parameter. Therefore, those that were not marked as visited (eligible) will remain as 0. Once this has completed, the current state $s$ is set to the next state $s'$ and the current action $a$ is set to the next action $a'$. The process, as embedded in the bot shooting logic, is illustrated in Fig. 4.

### C. Learning to Shoot

The success of any reinforcement learning algorithm relies on the design of suitable *states* (detailed descriptions of the current situation for the agent), *actions* (control statements for interaction with the environment) and *rewards* (positive or negative feedback for the agent). This section provides a detailed description of our design of the states, actions and rewards for the task of shooting. The state and action space for the current architecture was designed specifically for the map Training Day. This map was chosen due its small size and tendency to encourage almost constant combat between players. Since the reinforcement learning architecture is only concerned with shooting, the smaller map prevents players from having to excessively explore the map before encountering opponents. The architecture could be tailored to work equally and consistently between all maps by introducing very few changes. For instance, we discard the Z value when reading the relative velocity in the map Training Day, given the flat nature of the maps geometry. This, of course, has the consequence of ignoring some relevant velocity information when the opponent is jumping and dodging so in order for the state space to be more complete and representative of all maps, with complex geometry of varying sizes, this value would have to be reintroduced. Also, the discretized values for distance are specific to the Training Day map. Game logs were used to determine the min, max, and average distances of opponents during combat. These values were then used to create an approximate notion of "close," "medium," and "far" for the specific map. More generalized values would be required for the same distance categories to be applicable in all maps.

TABLE I
DISCRETIZED DISTANCE VALUES

| State | Distance | Player Widths |
|---|---|---|
| *Close* | 0 - 510 UU | 0 - 15 |
| *Medium* | 510 - 1700 UU | 15 - 50 |
| *Far* | >1700 UU | >50 |

TABLE II
DISCRETIZED SPEED VALUES

| State | Total Relative Velocity |
|---|---|
| *Regular* | 0 - 800 UU/sec |
| *Fast* | >800 UU/sec |

*1) States:* The state space is inspired by how humans perceive enemy players during FPS combat. While target selection is an interesting problem in itself, our implementation uses a simplified approach in which the bot will always engage in combat with the nearest visible player. We have taken into account its own system of measurement called unreal units (UU). These units are used when measuring distance, rotation and velocity. The collision cylinder of the NPC's graphical model is 34 units in diameter and 39 units in height. Each character in the game has an absolute location on the map represented by X, Y, and Z coordinates in UUs. The X and Y values are in the horizontal plane while the Z value represents the height of the character above a baseline. We measure the distance of the bot to the enemy and descretize these values into the ranges of "close," "medium," and "far" as detailed in Table I.

The bot will only determine the range of the current opponent of which it is engaging in combat. This will always be the closest visible player as mentioned earlier. The enemy is said to be close to the bot if the bot's current location falls inside a perimeter of 510 UUs units surrounding the enemy. To give the reader an idea of the size of this perimeter, it is equivalent to 15 player widths. The enemy is a medium distance from the bot if its location falls between 15 and 50 player widths from the bot and anything over 50 is considered far. The relative speed of the enemy can be "regular" or "fast" as shown in Table II.

These relative velocity state values were again determined based on log data and spectating games in progress and do not represent a universal notion of relative speed in the game. The values would need to be updated for other maps with the reintroduction of the Z value. As aforementioned, we only take the X and Y coordinates of the velocity vector into account when calculating relative velocity. The enemy's relative velocity to the bot is calculated and then the square root of the X value squared plus the Y value squared gives the total relative velocity. If this value falls below a certain threshold then the enemy is said to be moving at a regular speed. Anything above this threshold is treated as fast. The relative direction that the enemy is moving is also taken into account. The values for this state representation are shown in Table III. Three checks are carried out to determine how the enemy is moving. First, the enemy can be moving towards or away from the bot or not moving in either of those directions. Second, the enemy can also be moving either left or right or not moving in either of these directions. Third, the enemy can be jumping or not when moving in any direction and is stationary when not moving in any direction. In our definition of "stationary," the enemy can still be jumping on the spot.

TABLE III
DISCRETIZED MOVEMENT DIRECTION VALUES

| Check | Values |
|---|---|
| T/A Direction | Towards / Away / None |
| L/R Direction | Left / Right / None |
| Jumping | Yes / No |

TABLE IV
SHOOTING STATES

| Attributes | Number of values |
|---|---|
| Distance | 3 |
| Velocity | 2 |
| Jump | 2 |
| Direction | 9 |
| Rotation | 6 |
| Instant Hit | 2 |
| Total | 3*2*2*9*6*2 = **1296** |

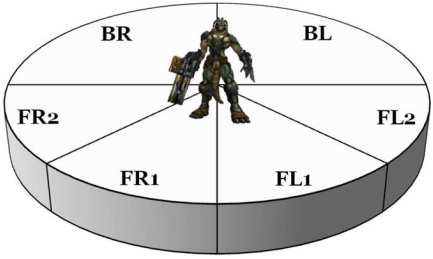

Fig. 5. Discretized values for the enemy's rotation.

There are 6 discrete values, shown in Fig. 5, for representing the direction in which the opponent is facing. These values are Front Right One (FR1), Front Right Two (FR2), Back Right (BR), Back Left (BL), Front Left Two (FL2), and Front Left One (FL1). The enemy will not always move in the same direction as it is facing but knowing which direction it is facing could be useful for anticipating the enemy's sequence of movements. The bot also takes into account whether the weapon they are using is an "instant hit" weapon or not. This means that there is no apparent delay from when the weapon is fired to hitting the target. Examples of such weapons are the sniper rifle and lightning gun which instantly hit their target once fired. Other guns shoot rockets, grenades and slow moving orbs which take time before hitting the target. The complete state space for shooting includes 1296 different states using the aforementioned checks. These are summarized in Table IV.

*2) Actions:* The actions that are available to the bot involve variations on how it shoots at enemy targets. We have identified six different categories of weapons to account for the variance in their functionality.

The *Instant Hit* category is for weapons that immediately hit where the cross-hairs are pointing once the trigger has been pulled. The primary mode[3] of the Sniper Rifle, Lightning Gun, and Shock Rifle all belong to this category. The Sniper Rifle and Lightning Gun don't have a shooting secondary mode but activate a zoomed in scope view for increased precision.

The primary mode of the Assault Rifle and both modes of the Mini Gun are examples of the *Machine Gun* category which spray a constant volley of bullets.

The *Projectile* category is made up of guns which shoot explosive projectiles. These include grenades from the secondary mode of the Assault Rifle and Flak Canon as well as an exploding paste from the Bio Rifle in primary mode. The secondary mode of the Bio Rifle is used for charging the weapon to produce a larger amount of paste.

*Slow Moving* guns, which shoot ammunition such as rockets or "orbs," involve a delay from when they are shot to when they reach the target. Examples of these guns include the secondary mode of the Shock Rifle and the primary mode of the Rocket Launcher and the Link Gun.

*Close Range* weapons are those that should be used when in close proximity to an opposing player. The Flak Canon is an example of this type of weapon which shoots a spread of flak (primary mode) that is very effective at close range. The Shield Gun, used as a last resort weapon for defense, causes a small amount of damage from close range in primary mode and acts as a shield deflecting enemy fire in secondary mode.

The final category of weapons *Other* includes all other weapons in the game that haven't been identified in one of the previous categories. The weapon categories are summarized in Table V.

Each category of gun has five actions associated with it in the current implementation. This results in 6480 state-action pairs for each category of gun or 38880 state-action pairs in total. The actions available for each gun are listed in Table VI.

The shooting actions for the bot involve receiving the planar coordinates of the enemy's location and then making slight adjustments to these or shooting directly at that area. The *Head*, *Mid*, and *Legs* actions take the *X*-axis and *Y*-axis values directly and the *Z*-axis value is set to head height, the midriff or the legs of the opponent respectively. Shooting *left* and *right* involves skewing the shooting in that direction by incrementing/decrementing the X value of the target location by a fixed number of UUs. In the current implementation, the amount of skew added comes from fixed values that are specific to the weapon type. A possible improvement for future implementations could involve dynamically determining this skew based on the relative velocity of the opponent and the nature of the weapon being used. The *Player* action uses the inbuilt targeting which takes an enemy player as an argument and continually shoots at that player, regardless of their movement. This is essentially "locking on" to the opponent but since actions are chosen multiple times a second by the reinforcement learner this shouldn't be apparent to the human opposition. Experienced human players can often be very accurate, just not constantly flawless. The *Location* action shoots directly at the exact location of the opponent. There are three variants of shooting above the opponent, (*Above*, *Above-2* and *Above-3*), which differ by the distance above the player with *Above-3* being the highest. These "Above" actions are designed so that the bot can find the correct height above the opponent so that the resulting trajectory will lead to causing damage. The further away the opponent is, the greater the height required in order for the aim to be successful. *Left-2* and *Right-2*, which are found in the *Slow* category, provide a bigger adjustment in each direction to account for the slow moving ammunition.

---

[3]All weapons have a primary and secondary mode activated by left and right mouse clicks, respectively.

TABLE V
THE DIFFERENT TYPES OF GUN AVAILABLE TO THE BOT

| Instant Hit | Machine Gun |
|---|---|
| Sniper Rifle (Primary (No Sec.)) | Assault Rifle (Primary) |
| Lightning Gun (Primary (No Sec.)) | Mini Gun (Primary) |
| Shock Rifle (Primary) | Mini Gun (Secondary) |
| **Projectile** | **Slow Moving** |
| Assault Rifle (Secondary) | Shock Rifle (Secondary) |
| Flak Canon (Secondary) | Rocket Launcher (Primary) |
| Bio Rifle (Primary (No Sec.)) | Link Gun (Primary) |
| **Close Range** | **Other** |
| Flak Canon (Primary) | |
| Shield Gun (Primary) | (All other guns picked up) |
| Shield Gun (Secondary) | |

TABLE VI
SHOOTING ACTIONS FOR SPECIFIC GUN TYPES

| Instant | Machine | Projectile | Slow | Close | Other |
|---|---|---|---|---|---|
| Head | Player | Player | Player | Head | Head |
| Mid | Location | Location | Left | Mid | Mid |
| Legs | Head | Above | Left-2 | Legs | Legs |
| Left | Left | Above-2 | Right | Left | Left |
| Right | Right | Above-3 | Right-2 | Right | Right |

Unlike previous work in the literature, our shooting mechanism is being refined over time as the bot learns with in-game experience. While there is some randomness present to enable exploring in the policy, the bot constantly adapts over time based on continuous feedback, similar to a human player. Human players constantly adapt and learn what works best and then try to reproduce these actions as often as they can. Mistakes are, of course, made from time to time which are being accounted for here with random action-selection occurring a percentage of the time during exploration. At the early stages of learning, the bot will not know the best actions to take so they are all equally likely to be chosen.

Weapon selection for the bot is taken from hard-coded priority tables of close, medium and far combat based on human knowledge of the weapon capabilities. These tables were inspired by a similar system in the $UT^2$ bot [17]. The bot will use the best available weapon that it has, according to these tables, based on the current distance from the opponent. Weapon selection in itself is a task that could be learned but our current research is focused on shooting technique so we opted to use human knowledge for weapon selection.

*3) Rewards:* The reward that the bot receives is dynamic and related directly to the amount of damage caused by the shooting action. The bot receives a small penalty of $-1$ if the action taken doesn't result in causing any damage to an opponent. This ensures that the bot is always striving towards the long term goal of causing the most damage that it can given the circumstances and minimizing unsuccessful shots that do not cause any damage.

D. Architecture Summary

To the best of the authors knowledge, this architecture is the first to use reinforcement learning to enable NPCs to learn and adapt the skill of shooting over time based on in-game experience. This approach is novel in that the bot will constantly adjust its shooting technique as it gathers knowledge of what works well and what does not. This approach is inspired by how humans learn to play these games. Modern FPS games have fast paced, complex environments that require instantaneous decision making. We have developed a state space and action space representation to facilitate the bots perception of opponents in the environment by reading key details. We have also tailored suitable actions to react in given circumstances. Reading important features of the opponents combat movements coupled with the "damage caused" reward signal are then used to drive real-time, knowledge-based decision making.

Traditional approaches to NPC shooting in FPS games involve limiting the ability of the bot by incorporating either a delay before shooting or purposefully missing the target to simulate lower ability from the bot. Examples of some of the restrictions imposed on bots are shown later in Section IV.A. The main drawback of this approach is the lack of adaptation.

Once a human player forms a strategy to beat such an opponent, there is no longer a challenge and the gameplay becomes highly predictable. Our approach constantly adapts the shooting technique based on in-game experience. In order to advance the state-of-the-art, we believe that computer-controlled opponents should adapt to their surroundings and improve with experience over time. Enabling learning in individual tasks, such as shooting, leads to less predictable and more engaging gameplay.

## IV. EXPERIMENTATION AND ANALYSIS

### A. Details

Three individual RL-Shooter bots were trained against native scripted opponents from the game with varying difficulty. These native bots ship with the game and each of them has a hard-coded scripted strategy that dictates how they behave. A discussion of these bots and a list of all the skill levels available can be found at the Liandri Archives: Beyond Unreal website[4]. In our experiments we use three skill levels, Novice, Experienced, and Adept:

- **Opponent Level 1 (Novice)**—60% movement speed, static during combat, poor accuracy (can be 30 degrees off target), 30 degrees field of view.
- **Opponent Level 3 (Experienced)**—80% movement speed, can strafe during combat, better accuracy and has faster turning, 40 degrees field of view.
- **Opponent Level 5 (Adept)**—Full speed, dodges incoming fire, tries to get closer during combat, 80 degrees field of view with even faster turning.

Each experiment run involved the RL-Shooter bot competing against three opponents that have the same skill level as each other, on the Training Day map in a series of thirty minute games. Training Day is a small map which encourages almost constant combat between opponents. We chose this map since the focus of our experimentation was on the shooting capabilities of the RL-Shooter bot. A total of three experiment runs took place, one for each of the opponent skill levels mentioned above. There was no score limit on the games and they only finished once the thirty minute time limit had elapsed. Each time the RL-Shooter bot was killed the state-action table was written out to a file. These files represent a "snap-shot" of the bots decision-making strategy for shooting at that moment in time. Each bot starts out with no knowledge (Q-table full of 0's) and then, as the bot gains more experience, the tables become more populated and include decisions for a wider variety of situations. The amount of exploration being carried out in the policy of the learners was dependent on the values from Table VII. For the first 10 000 lives the bot is randomly selecting an action half of the time. The other half of the time it is using knowledge that it has built up from experience (choosing the action with the greatest Q-Value based on previous rewards received). During exploration, we included a mechanism for choosing randomly from actions which haven't been selected in the past to maximize the total number of state-action value estimates that are produced. The exploration rate is reduced by 10% every ten thousand lives until is remains static at 5% once the bot has been killed over 50 000 times.

[4]http://liandri.beyondunreal.com/Bot.

TABLE VII
EXPLORATION RATE OF THE RL-SHOOTER BOT

| Lives | Exploration Rate |
|---|---|
| 0 - 9,999 | 50% |
| 10,000 - 19,999 | 40% |
| 20,000 - 29,999 | 30% |
| 30,000 - 39,999 | 20% |
| 40,000 - 49,999 | 10% |
| 50,000 or greater | 5% |

### B. RL-Shooter Bot 60 000 Lives: Results and Analysis

This section and the next one present the experimentation results from two different perspectives. In this section, we look at the different trends that occur with the bot having lived and died 60 000 times. This is followed in Section C by analyzing *the same results* from the perspective of the 30 minute games that were played, as opposed to the individual lives. The Level 5 skilled opponent had played 350 games as its death count approached 60 000. For this reason, our comparative game analysis of the three skill levels is carried out over 350 games.

In this section, we look at the results and statistics gathered from each of the RL-Shooter bots playing against opponents with different skill levels (Level 1, Level 3, and Level 5 opponents). From here on, we will refer to the RL-Shooter bot playing against Level 1 opponents as RL-Shooter-1 and the other two, playing against Level 3 and Level 5 opponents, as RL-Shooter-3 and RL-Shooter-5, respectively. We analyse the results of the bots having lived through 60 000 lives with a decreasing exploration rate as described in Table VII.

First, in Table VIII, we can see the total kills, deaths and suicides accumulated over the 60 000 lives for each bot. This table also shows the kill-death (KD) ratio which computes how many kills were achieved for each death (either by the other player or by suicide). RL-Shooter-1 has a KD ratio of 1.87:1 with almost 20% of its deaths coming from suicides. Suicides occur in the game when the bot uses an explosive weapon too close to an opponent or wall and can also occur if a bot falls into a lava pit. Although the Training Day map is small, there are three separate areas where bots can fall to their deaths. The RL-Shooter-3 bot appears to be more evenly matched with its opponents and has a KD ratio of 1.07:1. Deaths by suicide correspond to 12% of the bots overall deaths. The number of suicides appears to be directly linked to the number of kills which suggests that the majority of suicides are inflicted by the bot's own weapon as opposed to falling into a pit as mentioned earlier. This is confirmed further by the reduced suicide rate (10%) and kill totals for the RL-Shooter-5 bot. The RL-Shooter-5 has a negative KD ratio with 0.67 kills to every death.

Table IX shows the average and standard deviation of hits, misses, and reward received for the 60 000 lives. A *hit* is recorded each time the bot shoots its weapon and causes damage to an opponent. A *miss* is recorded when the weapon is fired but fails to cause any damage. The *reward* corresponds to the exact amount of damage inflicted on an opponent for the current hit or −1 if no damage resulted from firing the weapon. RL-Shooter-1 fires the most shots per life on average with 36.66 (27% hits; 73% misses). This would be expected as weaker opposition would afford the bot more time to be shooting, both accurately and inaccurately. The shots per life and accuracy decrease as the skill level of the opposition increases with RL-Shooter-3 shooting an average of 28.71 shots (25% hits; 75% misses) and RL-Shooter-5 shooting an average of 22.53 shots (21% hits; 79% misses).

TABLE VIII
TOTAL KILLS, DEATHS AND SUICIDES AND KILL-DEATH RATIO

| Opponent Skill Level | Total Kills | Total Deaths By Others | Total Deaths By Suicide | KD Ratio |
|---|---|---|---|---|
| Level 1 | 112420 | 48299 | 11701 | 1.87:1 |
| Level 3 | 63934 | 52994 | 7006 | 1.07:1 |
| Level 5 | 40466 | 54136 | 5864 | 0.67:1 |

TABLE IX
AVERAGE AND STANDARD DEVIATION VALUES AFTER 60 000 LIVES

| Opponent Skill Level | Hits Avg (Std Dev) | Misses Avg (Std Dev) | Reward Avg (Std Dev) |
|---|---|---|---|
| Level 1 | 9.82 (±7.42) | 26.84 (±18.46) | 79.82 (±83.00) |
| Level 3 | 7.30 (±5.69) | 21.41 (±11.68) | 47.71 (±50.38) |
| Level 5 | 4.70 (±3.79) | 17.83 (±8.73) | 25.06 (±33.37) |

TABLE X
PERCENTAGES OF HITS AND MISSES OVER THE 60 000 LIVES

| Opponent Skill Level | Hit Percentage | Miss Percentage |
|---|---|---|
| Level 1 | 27% | 73% |
| Level 3 | 25% | 75% |
| Level 5 | 21% | 79% |

TABLE XI
MINIMUM, MAXIMUM AND MEDIAN VALUES AFTER 60 000 LIVES

| Opponent Skill Level | Hits Min\|Max\|Med | Misses Min\|Max\|Med | Reward Min\|Max\|Med |
|---|---|---|---|
| Level 1 | 0 \| 72 \| 8 | 0 \| 232 \| 22 | -50 \| 1261 \| 56 |
| Level 3 | 0 \| 46 \| 6 | 0 \| 115 \| 19 | -50 \| 538 \| 35 |
| Level 5 | 0 \| 39 \| 4 | 0 \| 99 \| 17 | -49 \| 322 \| 16 |

TABLE XII
AVERAGES PER GAME AFTER 350 GAMES (30 MINUTE TIME LIMIT)

|  | Level 1 | Level 3 | Level 5 |
|---|---|---|---|
| Weapons Collected | 171.12 | 105.85 | 98.43 |
| Ammunition Collected | 72.54 | 57.74 | 57.36 |
| Time Moving (mins) | 20.81 | 20.66 | 20.30 |
| Distance Travelled (UT) | 489464.34 | 472513.43 | 459385.39 |

While the level of shooting inaccuracy may seem quite high for all of the bots, they are all still performing at a competitive standard as evidenced by Table VIII. It is important to remember that hits are only recorded when the bot is shooting and the system indicates that it is currently causing damage. All other shots are classified as misses. The actual damage caused by individual hits also varies greatly depending on the gun type used and the opponents proximity to explosive ammunition from certain guns.

Table XI lists the minimum, maximum and median values of the hits, misses and rewards over the 60 000 lives. The minimum numbers of hits and misses for each of the levels was zero. This is a result of the bot spawning into the map and being killed before it has a chance to fire its weapon. The maximum numbers of hits, misses and rewards are again closely dependent on the opposition skill level and the large difference between these and the median values shows the amount of variance from life to life.

The reward, as mentioned earlier, corresponds directly to the amount of damage that the bots successfully inflict on their opponents. The value is set to 0 at the beginning of each life and then accumulates based on any damage caused or is decremented by 1 for shots that do not result in any damage. The results for each of the skill levels do not show a clear upward trend for total reward per life received over time. There could be many reasons for this. The ammunition from the different guns that can be picked up from the map cause varying degrees of damage upon successfully hitting an opposing player. While the RL-Shooter bots are learning different strategies for each of the different types of gun, they have no control over which weapon they have available to them during each life. They are prioritizing the use of the more powerful weapons when they are available but during many lives, as evidenced by the shooting time average data from Table XII, they have not acquired these more powerful weapons. The small number of actions available for each gun type could also be a reason behind performing well in the earlier games even when selecting randomly. On some occasions, the bot received a substantial total reward during its lifetime but it is inconclusive as to whether this was occurring randomly, given the nature of the game (real time, multiple opponents, small map etc.), or whether the bots were improving their action selection as they experienced new situations and then took advantage of this knowledge when these situations occurred at a later stage.

*C. Thirty Minute Games: Results and Analysis*

This section analyzes the results and statistics based on individual games as opposed to the lives of the bots which we looked at in the previous section. Specifically, we look at 350 games, each with a duration of 30 min, for the three different opponent skill levels. All of the following results and statistics are reviewed on a "per game" basis.

Table XII lists some game statistics averaged over the 350 games. RL-Shooter-1 collected nearly twice as many weapons on average as the other two bots. All players in the game start each life with an Assault Rifle and must pick up additional weapons and ammunition from different points around the map. The Assault Rifle is a weak weapon and is only used when a better weapon is not available. Playing against lesser opposition gives the bot many more opportunities to pick up different weapons and also replenish their ammunition with pick-ups. All three bots spent the same amount of time moving which was just over 20 min. This would be expected as they were all using the same navigation modules which did not include any learning. Time spent not moving would include the short delays between when the bot is killed and when it spawns back to life on the map. The average distance travelled for each bot over the 350 games is also shown and this is measured in UUs. RL-Shooter-1 travels 16951 UUs more than RL-Shooter-3 per match, on average, while RL-Shooter-3 travels 13128 UUs more than RL-Shooter-5. This would suggest that as the skill level of the opponents increase the bots have less opportunity to traverse the map and thus miss out on important pick-ups.

TABLE XIII
SHOOTING TIME AVERAGES AFTER 350 GAMES (30 MINUTE TIME LIMIT)

|  | Level 1 | Level 3 | Level 5 |
|---|---|---|---|
| *Total Time Shooting (mins)* | 16.67 | 17.10 | 15.68 |
| *Shooting Assault Rifle (mins)* | 11.07 | 14.25 | 13.88 |
| *Shooting Shock Rifle (mins)* | 2.24 | 1.413 | 0.85 |
| *Shooting Flak Canon (mins)* | 1.92 | 0.74 | 0.54 |
| *Shooting Mini Gun (mins)* | 1.39 | 0.67 | 0.39 |
| *Shooting Shield Gun (mins)* | 0.03 | 0.01 | 0.002 |

TABLE XIV
AVERAGE AND STANDARD DEVIATION VALUES AFTER 350 GAMES

| Opponent Skill Level | Killed Avg (Std Dev) | Killed By Avg (Std Dev) | Suicides Avg (Std Dev) |
|---|---|---|---|
| *Level 1* | 251.08 (±4.71) | 100.94 (±14.36) | 24.75 (±8.46) |
| *Level 3* | 178.02 (±11.51) | 140.33 (±7.73) | 18.78 (±4.61) |
| *Level 5* | 135.06 (±10.61) | 168.73 (±9.38) | 18.54 (±4.34) |

TABLE XV
MINIMUM, MAXIMUM AND DIFFERENCE VALUES AFTER 350 GAMES

| Opponent Skill Level | Killed Min\|Max\|Diff | Killed By Min\|Max\|Diff | Suicides Min\|Max\|Diff |
|---|---|---|---|
| *Level 1* | 177 \| 287 \| 110 | 78 \| 121 \| 43 | 12 \| 43 \| 31 |
| *Level 3* | 139 \| 213 \| 74 | 118 \| 167 \| 49 | 8 \| 36 \| 28 |
| *Level 5* | 89 \| 158 \| 69 | 125 \| 190 \| 65 | 3 \| 33 \| 30 |

Table XIII shows the average amount of time shooting (in minutes) per game and also lists the shooting time for each of the individual guns. From the table, we can see that RL-Shooter-3 spends the most time shooting on average and also spends the most time using the Assault Rifle. RL-Shooter-1 does not use this default gun as much as the other bots because it is able to pick up stronger weapons from the map. The Shield Gun, which the bots also spawn with, is seldom used in any case as this is a "last resort" weapon which helps the bot to defend itself while searching for a more effective weapon. The small map with multiple opponents meant that the bots rarely got into a situation where the Shield Gun was the only remaining option.

Table XIV shows the average kills, deaths by others (Killed By) and suicides from the 350 games. One fifth of the deaths to the RL-Shooter-1 bot were self-inflicted. Aside from this fact, the bot managed to keep an impressive 2:1 kill-death ratio. RL-Shooter-3 bot was more closely matched with its opponents (1.12:1 KD) where as RL-Shooter-5 bot had a negative kill-death ratio of 0.72:1. RL-Shooter-3 and RL-Shooter-5 had very similar suicide rates to each other.

The minimum, maximum and difference (between min and max) of Kills, Killed By, and Suicides after the 350 games are shown in Table XV. This table gives an idea of the range of variance between games when playing against each of the skill levels.

Another indicator of performance in FPS games is known as a Kill Streak. This is a record of the total amount of kills that a bot can make without dying. The maximum Kill Streak was recorded for each of the games and is shown in Fig. 6. The highest Kill Streak per game for RL-Shooter-1 usually falls between 7 and 10 for each game. This appears to change, however, as more shooting experience is acquired and falls between 11 and 16 on many occasions reaching even as high as 20. RL-Shooter-3 usually achieves maximum Kill Streaks of either 5 or 6 but again these increase over time with the highest that it reaches being 11. RL-Shooter-5 is less successful at achieving high Kill Streaks with the majority of them being either 3 or 4. It does, however, manage to achieve a Kill Streak of 9 on two occasions.

Fig. 7 shows the total number of kills that the RL-Shooter bots achieved in each of the 350 games. A clear separation of the results can be seen from the graph. RL-Shooter-1 manages to kill opponents in the range of 200 to 300 times each game. This range drops to between 150 and 200 for RL-Shooter-3 and again drops to falling mostly between 100 and 150 for RL-Shooter-5. Improvements in performance over time, while not significant, are more evident against the Level 3 and Level 5 opponents. This would suggest that the RL-Shooter-1 bot learns the best strategy to use against the weaker opponent at an early stage and then only ever matches this, at best, in the subsequent games.

The total number of deaths from the same 350 games are shown in Fig. 8. There is once again a clear separation of the data based on the skill level. While the number of deaths of the RL-Shooter-5 bot mostly fall between 160 and 180, there are a number of occasions midway through the games in which they fall within the range expected of a Level 3 bot (120 to 160). The number of deaths for the RL-Shooter-1 bot are quite evenly dispersed between 80 and 120 throughout all of the games.

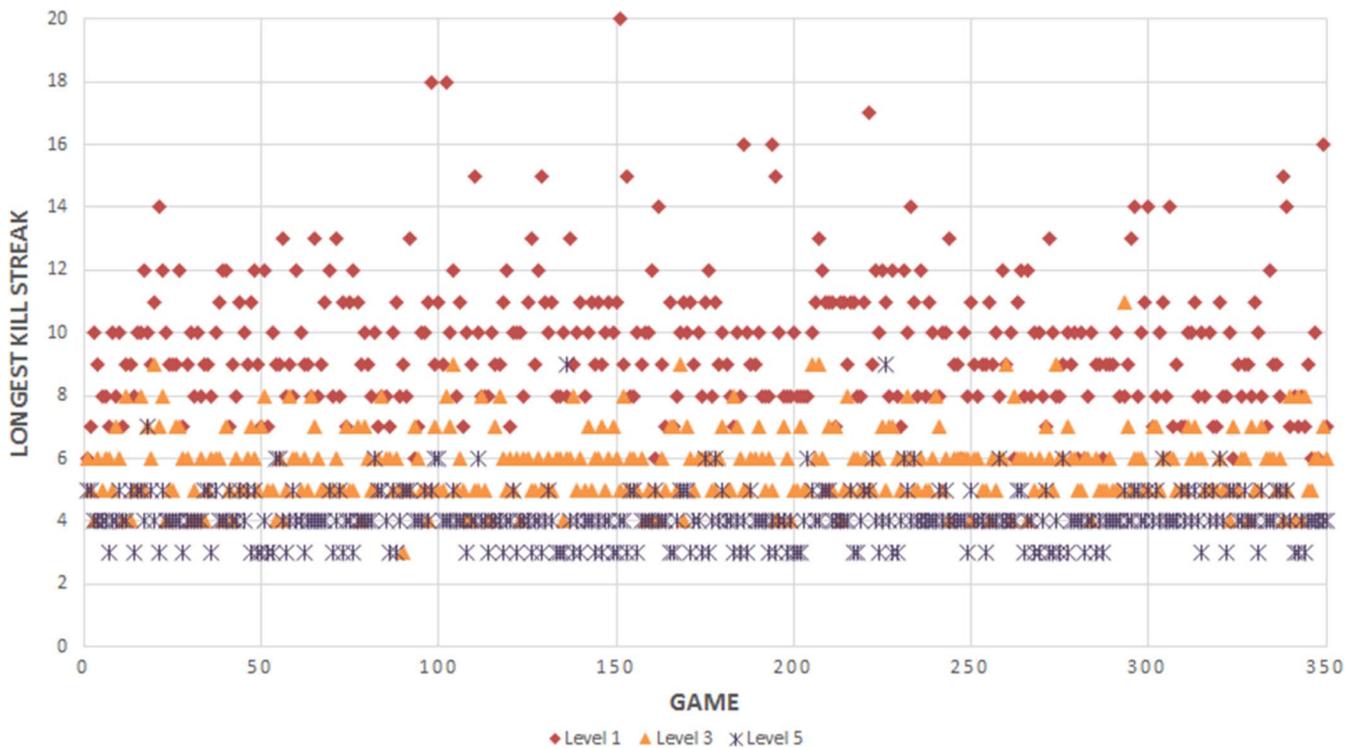

Fig. 6. Longest kill streak per game for each of the opponent skill levels.

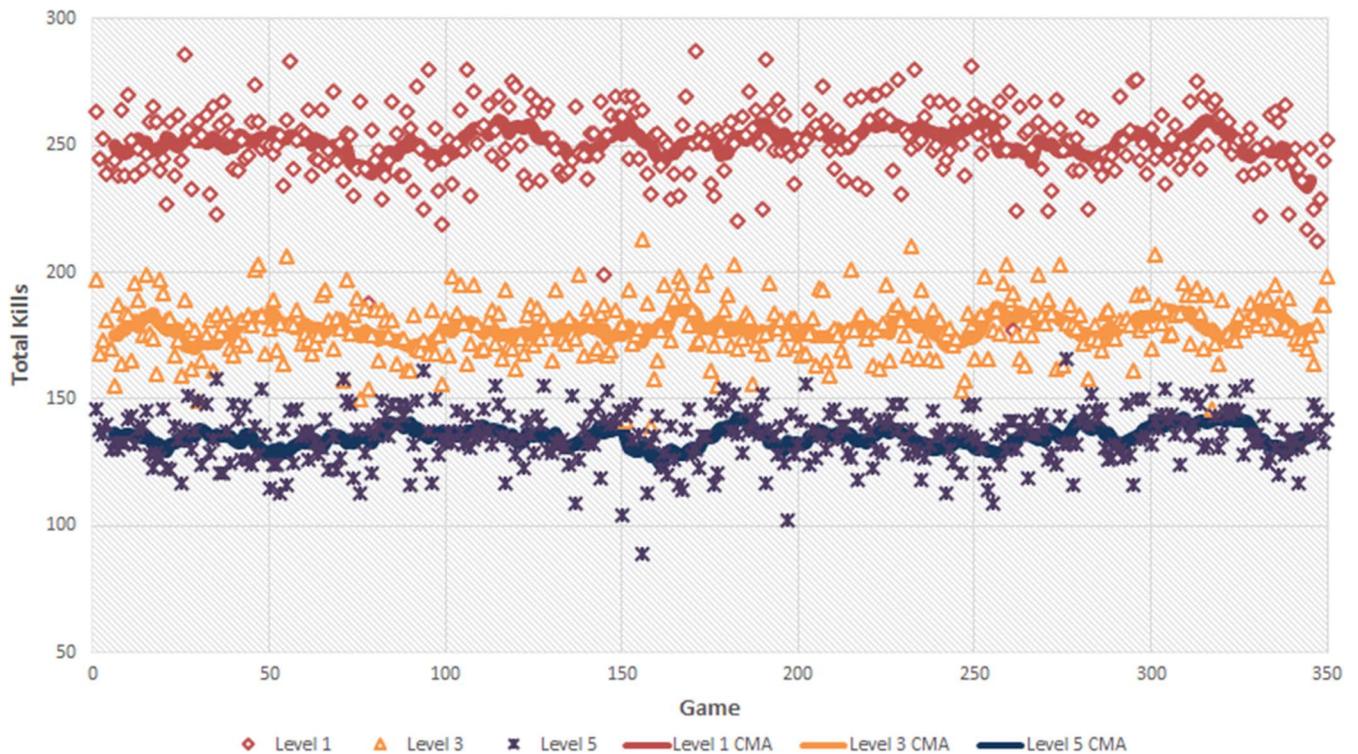

Fig. 7. Total number of kills per game and Centred Moving Average of kills for each of the opponent skill levels.

In order to investigate the presence of any trends in the data, Figs. 7 and 8 also show the Centred Moving Average (CMA) of the total kills and deaths, respectively. We use an 11-point sliding window for the CMA, so each point on the graph represents the average of the 11 samples on which it is centred. RL-Shooter-1 is the most consistent when it comes to kills in the early games. It appears to be gradually increasing the number of kills until a dip in performance around Game 80. It then slowly begins to recover before the total kills begin to fluctuate up and down. It is just beginning to recover from another dip in perfor-mance in the final games. The other two bots RL-Shooter-3 and RL-Shooter-5 show a similar fluctuating pattern with total kills. There appears to be little evidence to suggest that the total kills are improving consistently over time. This can also be said of the total deaths which show a similar amount of variance. We attribute this to the fact that the bots are choosing from a small subset of actions at each time step. The bot can be successful when randomly choosing from these actions. Although the best actions will not become apparent until the bots have built up experience, they may choose successful actions at an early stage given their limited choices.

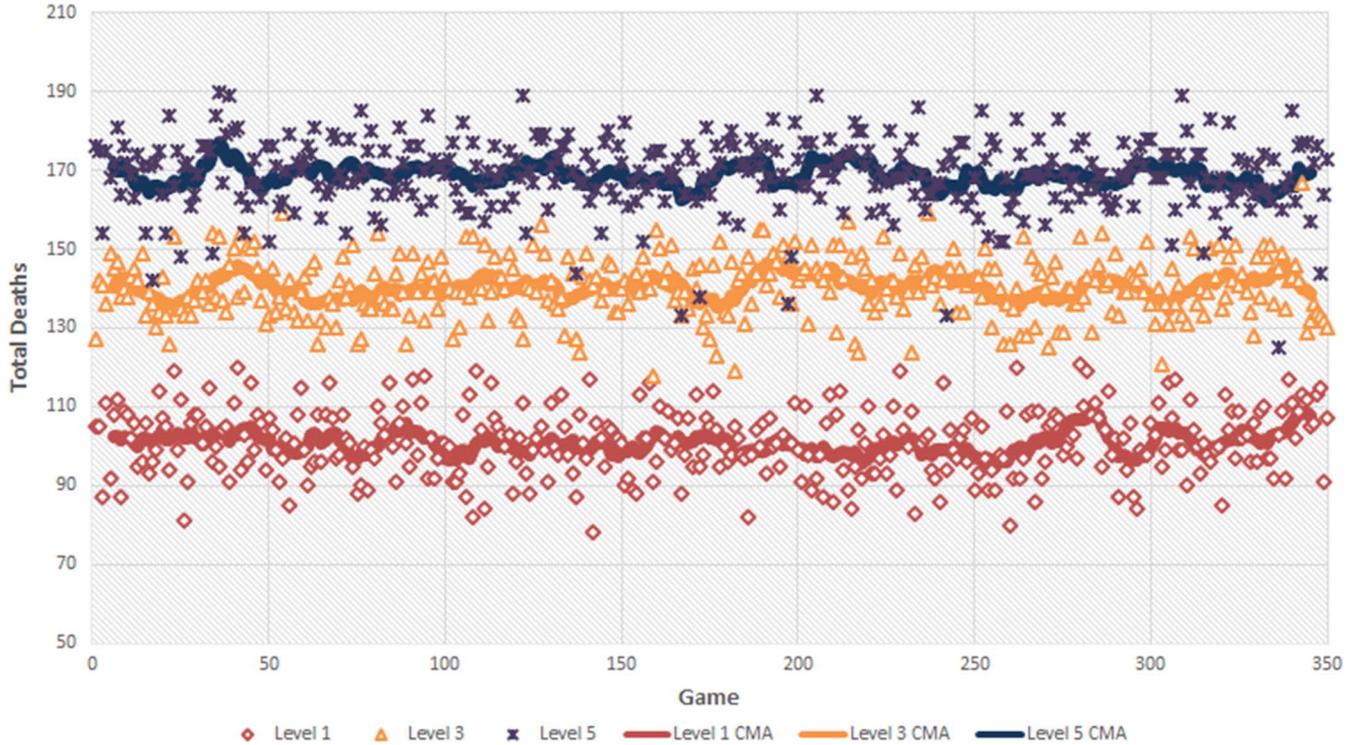

Fig. 8. Total number of deaths per game and Centred Moving Average of deaths for each of the opponent skill levels.

## V. Conclusion

This paper has described an architecture for enabling NPCs in FPS games to adapt their shooting technique using the Sarsa($\lambda$) reinforcement learning algorithm. The amount of damage caused to the opponent is read from the system and this dynamic value is used as the reward for shooting. Six categories of weapon were identified and, in the current implementation, the bot has a choice of five hand-crafted actions for each. The bot reads the current situation that it finds itself in from the system and then makes an informed decision, based on past experience, as to what the best shooting action is. The bot will continually adapt its decision-making with the long term objective of inflicting the most damage possible to opponents in the game.

In order to evaluate the reinforcement learning shooting architecture, we have carried out extensive experimentation by deploying it against native fixed-strategy opponent bots with different skill levels. The reason for pitching our bot against scripted opponents was to ensure that all of the games were played against opponents of a set skill level to facilitate a direct comparative analysis and make it easier to detect any possible trends in performance. This would be much more difficult to achieve with human opponents given the inherent variance in human game play and the amount of time that would be needed to run all of the games (with the same human players). That being said, we will move on to experimentation involving human opposition after further developing the system.

Reviewing the overall results that are presented in the preceding sections, the main trends that can be observed are:

- The RL-Shooter bots are able to perform at about the same level as the "Experienced" opponent, as was described above in Section IV-A; for example, its kill-death ratio against Level 3 opponents is approximately 1:1.
- When pitched against weaker opponents, the RL-Shooter bots perform better and when pitched against stronger opponents they perform worse; this can be seen in all of the results presented.
- From Figs. 7 and 8, there is not a clear pattern of the RL-Shooter bots improving in performance over time.

These results indicate how challenging it is for a bot with its relatively limited perception abilities and narrow range of actions to improve its performance over time. In our continuing work on this research topic, we will focus on identifying mechanisms by which we can improve the ability of the bots to demonstrate learning, by reviewing and refining our state representations, action representations, and reward design.

The overall aim of our research is to eventually generate bots that can compete with, and adapt to, human players and remove the predictability generally associated with computer-controlled opponents. The framework described in this paper is a platform that can be used by other researchers to tackle similar tasks. The results presented here are a comprehensive baseline against which future improvements can be measured.

ACKNOWLEDGMENT

The authors would like to thank the developers of the Pogamut 3 toolkit for providing invaluable technical support and advice during development.

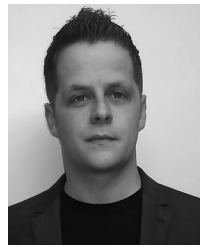

**Frank G. Glavin** was born in Galway, Ireland, on February 7, 1985. He received an honours degree in information technology from NUI Galway in 2006. He received the M.Sc. degree in applied computing and information technology from NUI Galway in 2010. This work involved developing a One-Sided Classification toolkit and carrying out experimentation on spectroscopy data.

He is currently a Ph.D. candidate researching the application of artificial intelligence techniques in modern computer games.

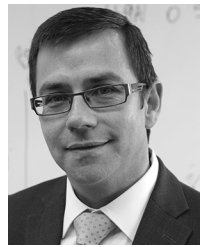

**Michael G. Madden** received the B.E. degree from NUI Galway in 1991.

He is the Head of the Information Technology Discipline and a Senior Lecturer with the National University of Ireland Galway, which he joined in 2000. He began his research career by working as a Ph.D. research assistant in Galway, then worked in professional R&D from 1995 to 2000. He has more than 80 publications, three patent filings, and cofounded a spin-out company based on his research.